\documentclass[letterpaper, 10 pt, conference]{ieeeconf}  

\IEEEoverridecommandlockouts                              

\usepackage{graphics} 
\usepackage{epsfig} 
\usepackage{mathptmx} 
\usepackage{times} 
\usepackage{amsmath} 
\usepackage{amssymb}  
\usepackage{subfig}
\usepackage{siunitx}
\usepackage{textcomp}
\usepackage{hyperref}
\usepackage[english]{babel}
\usepackage[font=small,labelfont=bf]{caption}
\newcommand{\g}{ g\textsubscript{0}}
\usepackage[backend=biber,      
    bibstyle=ieee,              
    citestyle=numeric-comp,     
    sortcites=true,   
    giveninits=true,            
    maxbibnames=3
]{biblatex}
\title{\LARGE \bf
Shaping in Practice: Training Wheels to Learn Fast Hopping Directly in Hardware
}

\author{Steve Heim, Felix Ruppert, Alborz A. Sarvestani, Alexander Spr{\"o}witz\\ Dynamic Locomotion Group \\ Max Planck Institute for Intelligent Systems, Germany
\thanks{\noindent contact: lastname@is.mpg.de}
\thanks{\noindent Dynamic Locomotion Group, Max Planck Institute for Intelligent Systems, Heisenbergstr. 3, 70569 Stuttgart, Germany}%
}

\begin{document}

\def\ps@IEEEtitlepagestyle{%
  \def\@oddfoot{\mycopyrightnotice}%
  \def\@evenfoot{}%
}
\def\mycopyrightnotice{%
  {\footnotesize © 2018 IEEE.Personal use of this material is permitted. Permission from IEEE must be obtained for all other uses, in any current or future media, including reprinting/republishing this material for advertising or promotional purposes,creating new collective works, for resale or redistribution to servers or lists, or reuse of any copyrighted component of this work in other works.\hfill}
  \gdef\mycopyrightnotice{}
}

\maketitle
\pagestyle{empty}

\begin{abstract}
Learning instead of designing robot controllers can greatly reduce engineering effort required, while also emphasizing robustness. Despite considerable progress in simulation, applying learning directly in hardware is still challenging, in part due to the necessity to explore potentially unstable parameters. We explore the concept of shaping the reward landscape with \emph{training wheels}; temporary modifications of the physical hardware that facilitate learning. We demonstrate the concept with a robot leg mounted on a boom learning to hop fast. This proof of concept embodies typical challenges such as instability and contact, while being simple enough to empirically map out and visualize the reward landscape. Based on our results we propose three criteria for designing effective training wheels for learning in robotics. A video synopsis can be found at \url{https://youtu.be/6iH5E3LrYh8}.
\end{abstract}

\section{INTRODUCTION}

In nature, animals learn to move with a grace and agility that is the envy of robotics engineers. One major challenge is that most algorithms rely on accurate models, which in turn also take a lot of engineering effort. Alternatively, reinforcement learning (RL) is a powerful paradigm that can work both model-based or \emph{model-free}. In addition, reinforcement learning is often able to learn from generic and even highly delayed reward signals: for example a legged robot might receive a reward for reaching a specific target location within a set time limit, and no reward for getting progressively closer. This allows for easy and intuitive assignment of rewards without constraining the behavior for achieving the goal. Despite these attractive features and promising achievements in simulation \cite{peng2016terrain}\cite{lillicrap2015continuous}, applying RL directly in hardware has proven challenging \cite{abbeel2007application}\cite{peters2008reinforcement}\cite{kober2013reinforcement} with only a handful of successes that actually run model-free \cite{tedrake2005learning}\cite{kohl2004policy}. \\
\begin{figure}[th]
    \centering
    \smallskip 
    \includegraphics[width=0.95\columnwidth]{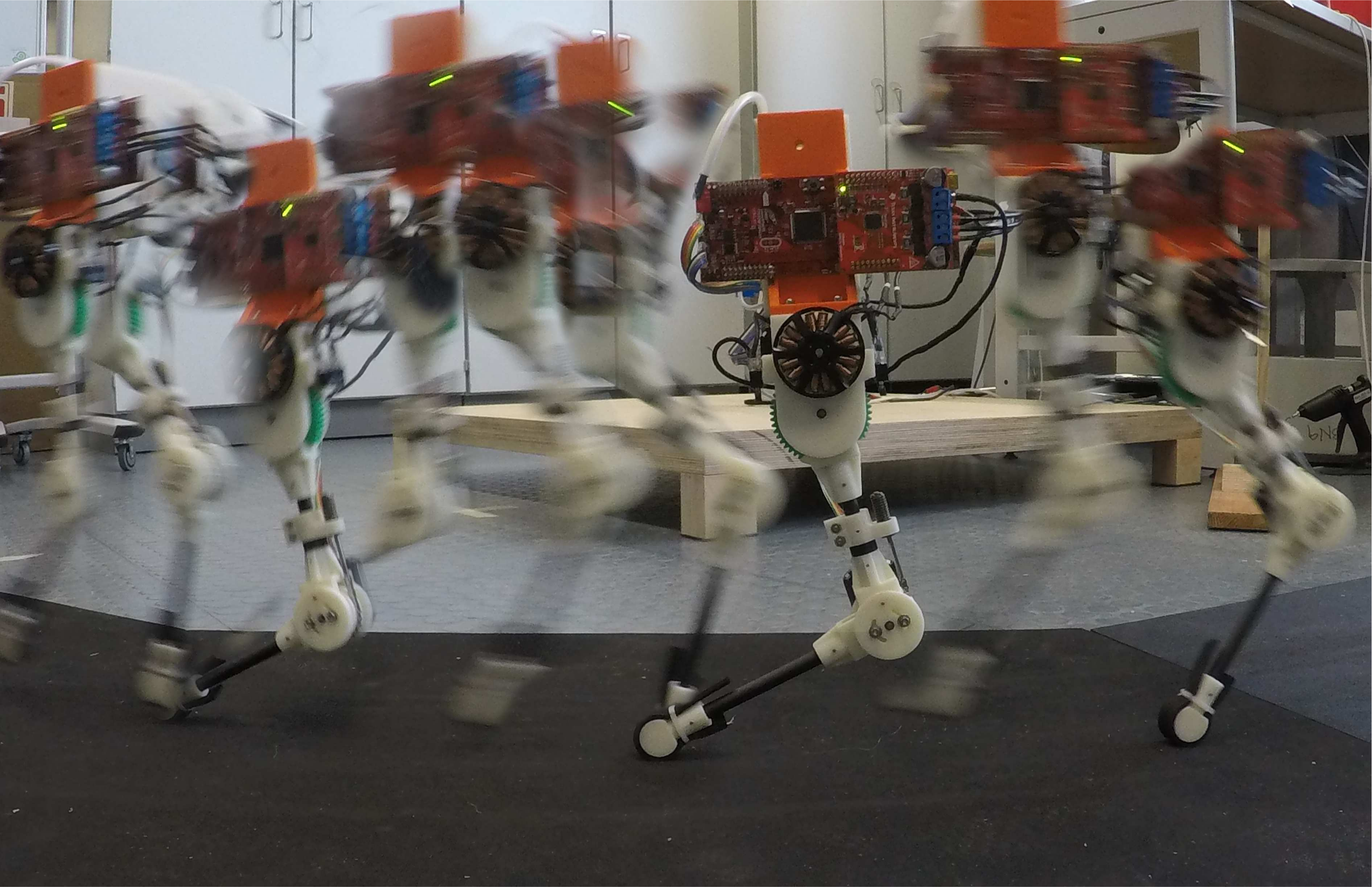}
    \caption{A multiple-exposure image of our robotic leg mounted on a boom hopping, showing the distinct stance and flight phases.}
    \label{fig:multiexp}
\end{figure}
One major challenge in hardware comes from the necessity to \emph{explore} the reward landscape. The landscape is usually non-convex, and often only subsets represent behaviors that actually accumulate reward: the rest of the landscape often looks flat, representing different behaviors that all receive the same or even no reward. Sampling from these regions provides no gradient information for the robot to learn from. This is particularly true when the reward is generic and delayed such as in the previous example: a policy that causes the robot to fall over immediately would get the same reward of zero as a policy that hops in place, even though the second policy arguably solves part of the locomotion problem \cite{raibert1986legged}. \\
This problem is even more accentuated in robots that are unstable, since instabilities often quickly lead to direct failure states. These failures generally lead to no reward, and can also damage the hardware. In practice, exploration is executed cautiously, usually locally. This combination means that large parts of the reward landscape are flat, and there is no salient gradient to lead the learning agent in the correct direction. \\
The exploration challenge can be solved by choosing a more appropriate policy parameterization, or with different exploration strategies such as intrinsic motivation \cite{chentanez2005intrinsically}. This can however be difficult to find and does not eliminate the flat regions, or the potentially damaging failures. \\
Another approach is to shape the reward landscape. A common method of shaping is to encode more information of the task in the reward \cite{gullapalli1992shaping}. The drawback is that it requires more prior knowledge of the task, and goes against the attractiveness of being able to choose rewards based on achieving a task rather than specifying a behavior\footnote{Whether the robot crawls, walks or runs should depend on the context of the situation and not on the goal.}. 
It is also possible to shape the landscape by proper mechanical design. For example, walking robots designed after passive-dynamic walkers \cite{mcgeer1990passive} have good stability properties for a wide range of policy parameters, allowing quick and reliable learning from even poor initializations \cite{tedrake2005learning}. The drawback is that designing the system around one specific behavior can be limiting in terms of versatility and design options. \\
We build on these ideas with the concept of \emph{training wheels}: shaping the landscape with \emph{temporary mechanical modifications of the robot that allow for easier learning}. To the best of our knowledge, this concept has only been briefly explored in simulation, even though initial results showed promise \cite{randlov2000shaping}. We present a proof of concept directly in hardware, applied to learning fast hopping of a monoped robot with a rolling foot: an underactuated, unstable system featuring hybrid dynamics. \\
We would like to note that we largely use the terminology of the RL community. In particular, the term \emph{environment} signifies everything that is beyond direct control of the learning agent. Take for example an agent whose policy outputs a desired joint position; then the environment includes not only the physical world the robot moves in, but also the robot itself and the PD motor controller used to track the desired joint position. For a more thorough treatment of RL see \cite{kober2013reinforcement}\cite{sutton1998reinforcement}.

\section{SETUP: MECHANICS, POLICY AND LEARNING SCHEME}
\begin{figure}[bthp]
    \centering
    \includegraphics[width=\columnwidth]{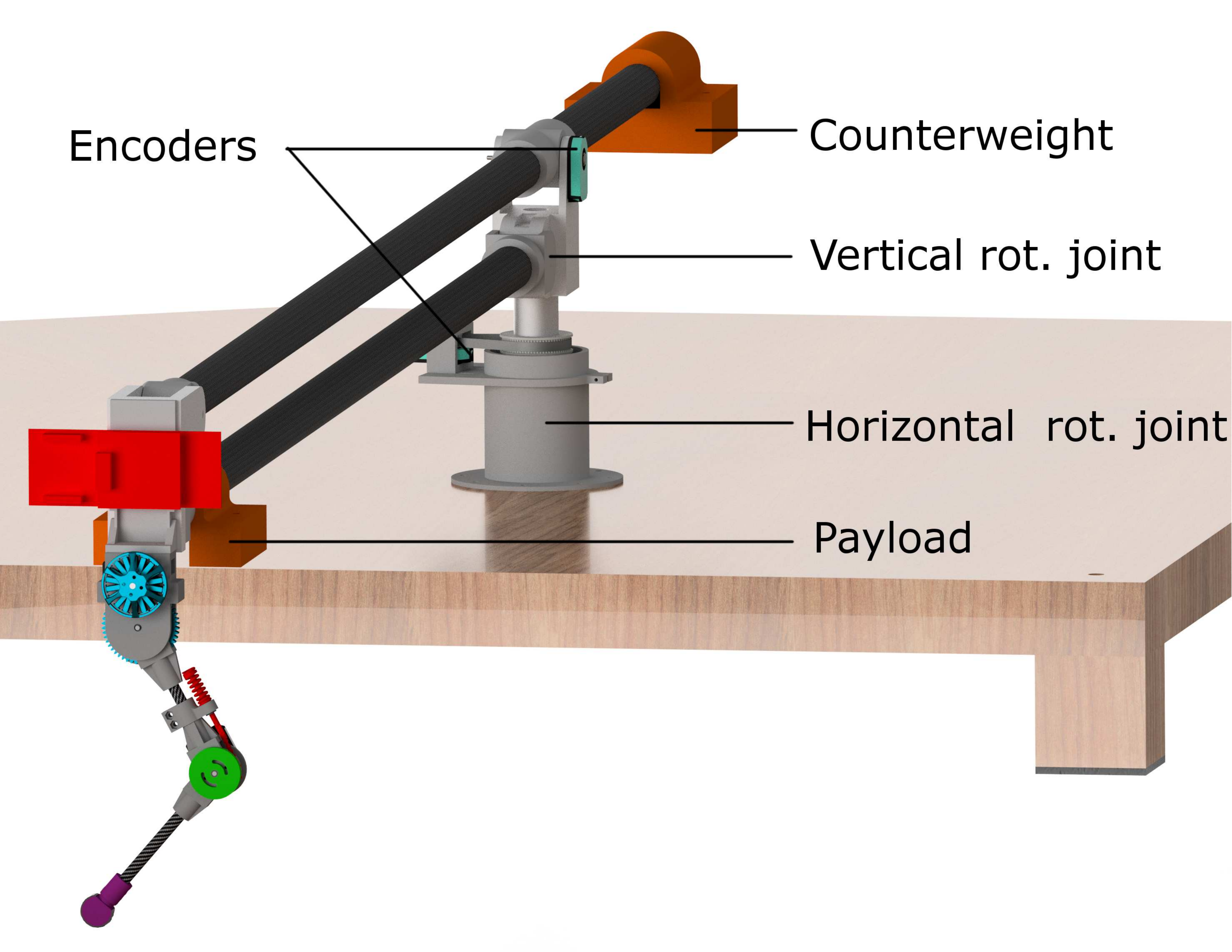}
    \caption{The entire robot consists of a leg mounted on a boom, with a total of four degrees of freedom. The counterweight balances out the mass of the boom without the leg or payload. The payload represents the mass of the batteries and additional electronics, which are offloaded via a tether.}
    \label{fig:robot}
\end{figure}

\begin{figure}[bthp]
    \centering
    \smallskip 
    \includegraphics[width=0.6\columnwidth]{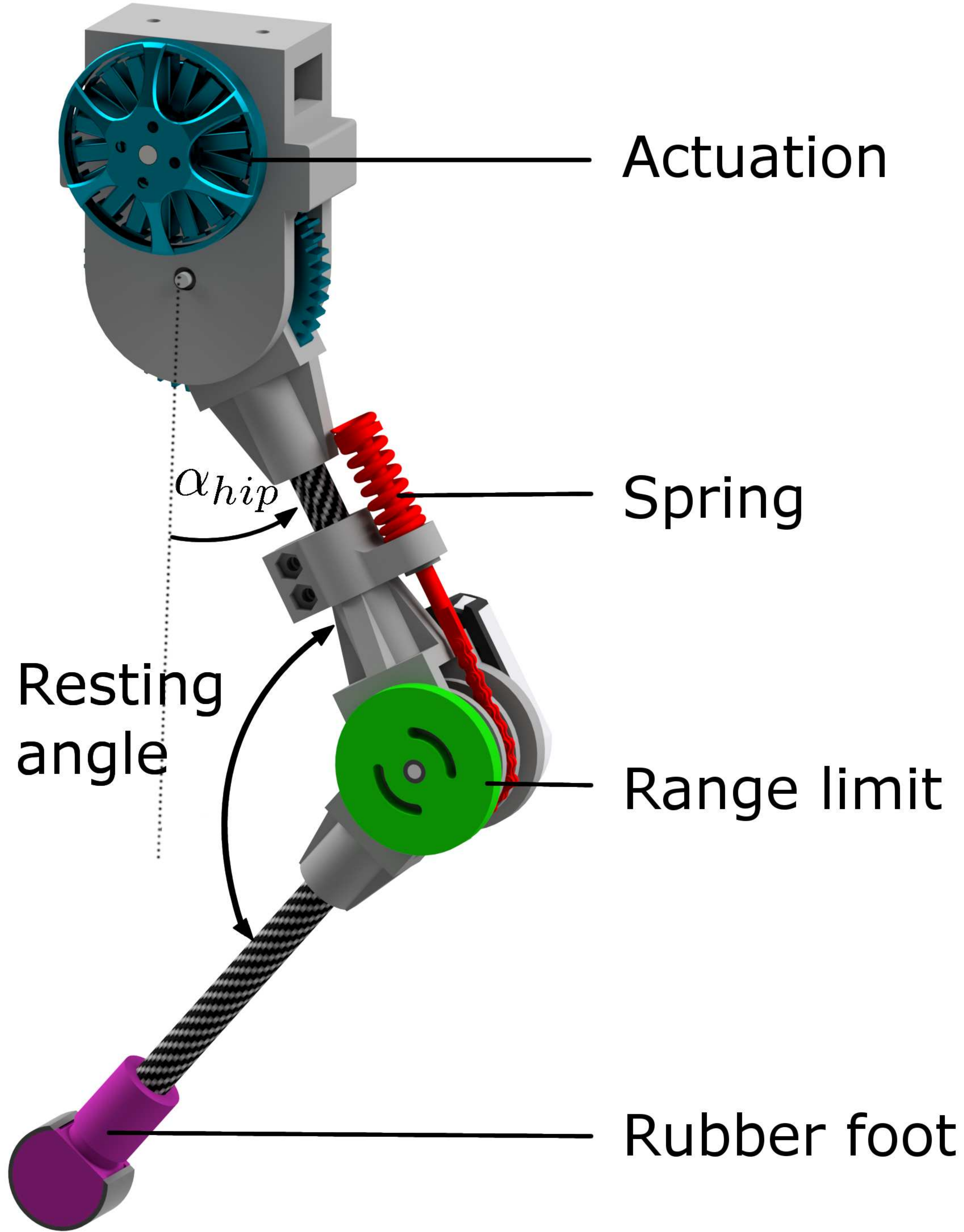}
    \caption{The two-segment leg has a brushless DC motor at the hip, and a passive compliant ankle joint. The spring is mounted to a cam mechanism, and the joint itself is limited in extension range.}
    \label{fig:leg}
\end{figure}
\begin{figure*}[t!]
    \centering
    \includegraphics[width=1\textwidth]{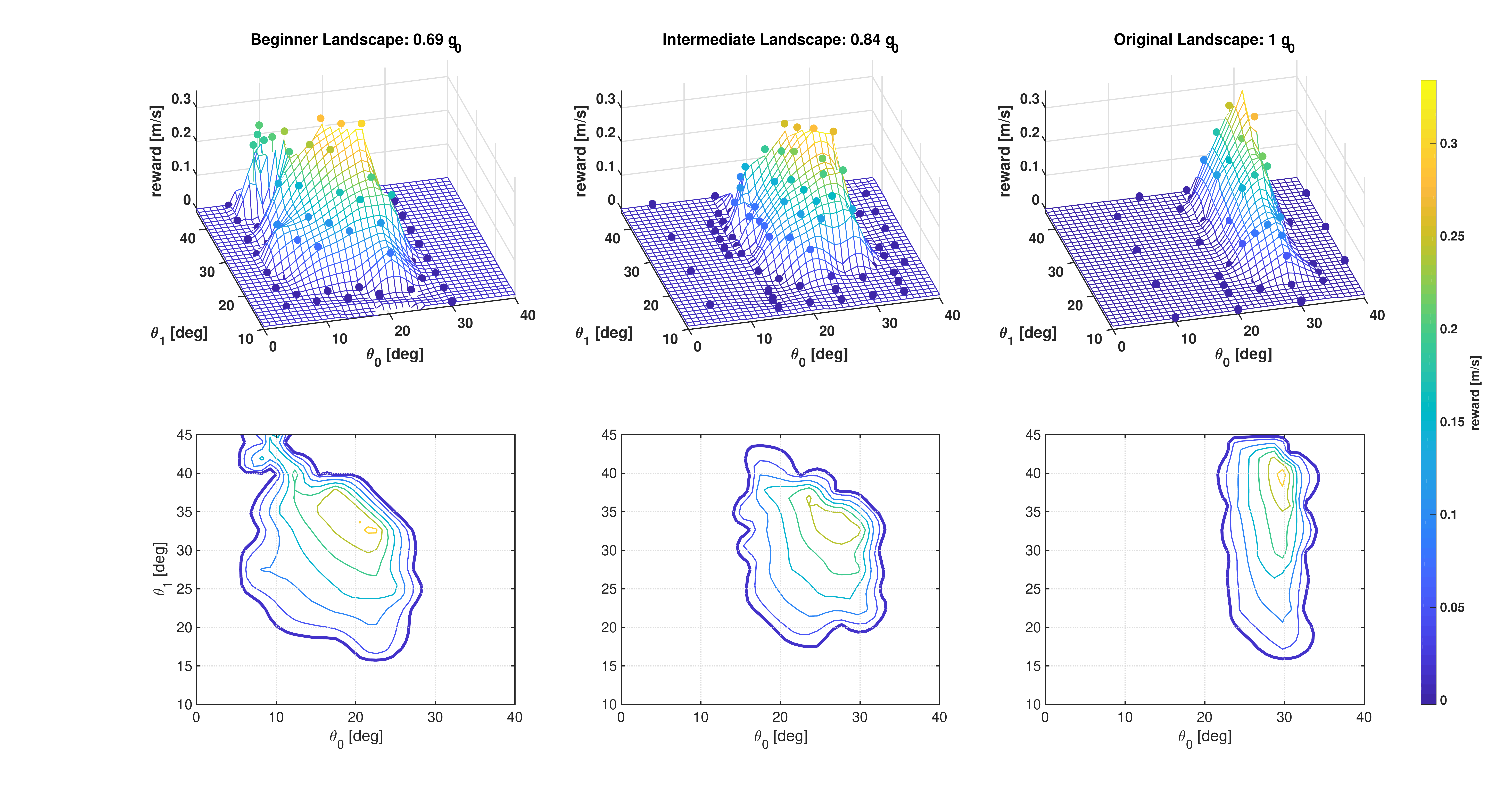}
    \caption{The landscapes of the beginner, intermediate  and original environments are visualized here. The upper row shows the sampled points (circles) and the resulting interpolated mesh, slightly offset for visibility. The more gradual climb in the lower gravity environments is visible. The contour maps in the second column more clearly show the change in shape of the `reward mountain', the shape of of the cliff and most importantly, the size of the basin of attraction for the learning system. The outer contour showing the set of parameters which can provide a gradient is outlined with a thicker line. If the learning agent only samples outside this set, it will not be able to accurately estimate a gradient.}
    \label{fig:3landscapes}
\end{figure*}

Our robot platform (Fig. \ref{fig:robot}) consists of a two-segment leg with a passive compliant ankle joint and an actuated hip joint, mounted on a boom which constrains the body to motion on a 2D surface. The robot thus has four degrees of freedom (DoFs) and a single actuator. The passive compliance at the ankle joint (Fig. \ref{fig:leg}) results in favorable natural dynamics \cite{rummel2008stable}, though the system is still passively unstable. \\
The learning task is to achieve fast hopping, and the reward for each rollout is the average speed with one additional condition: potentially damaging behavior, such as landing on the ankle instead of the foot, is tagged as a failure and receives no reward.  
The training wheels for this proof of concept are a simple change of the total mass of the robot body: essentially we allow learning in a reduced gravity environment.  \\
We choose these training wheels for two reasons: first, they should make it easier for the agent to learn the task. Second, they should also be easy to apply in practice, so the final behavior can be achieved with less engineering effort. Based on this we introduce our first criterion in designing training wheels: \emph{how easy are training wheels to apply to a generic set of robots}. For example, the weight of batteries, computation or other payload can easily be offloaded during an initial training phase for most robot designs. \\
We choose a simple policy with a 2D parameter space: the hip actuator tracks an open-loop sinusoidal position trajectory as follows

$$
\alpha_{hip} = \theta_0 + \theta_1\sin(\omega t) \eqno{(1)}
$$
where $\alpha_{hip}$ is the angular position of the hip, $\omega$ is a hardcoded angular frequency while $\theta_0$ and $\theta_1$, offset and amplitude parameters respectively, form the parameter space of the policy. This simple policy parameterization serves two purposes: first, a low-dimensional deterministic policy is amenable to the simplest of learning schemes, and thus eliminates the ambiguity of whether the training wheels or the algorithm implementation are responsible for the change in performance. In the results presented we choose $\omega = 9 \text{Hz}$, based on experience. Higher values achieve higher performance, but failures are also more violent and prone to damaging hardware. Since we also need to sample failure parameters to map out the landscape, we compromise between safety and performance. \\
We use stochastic gradient descent based on simple finite-difference methods \cite{peters2006policy}. More importantly, the low dimensionality allows us to empirically map out and inspect the landscape of the learning problem as a 3D surface as seen in Fig. \ref{fig:3landscapes}. This allows us to compare the landscapes with and without training wheels in detail, and show the change in learning performance across each landscape. 

\subsection{Hardware Details}
Each DoF of the boom and leg is measured with a rotary encoder (CUI ATM102-V). The boom arm has a length of $1.5 [m]$ from pivot to the leg, and is counterweighted to completely offset its own mass without the leg. The ankle joint of the leg (Fig. \ref{fig:leg}) is mechanically limited to $130$\textdegree{} in one direction, and has a spring with a stiffness of $6\ [\frac{N}{mm}]$ attached to a cam mechanism with a radius of $15\ [mm]$. This spring is slightly preloaded such that it always returns to the resting angle of $130$\textdegree{}. The upper and lower leg segments measure $110\ [mm]$ and $136\ [mm]$ respectively, and the virtual leg length from hip to foot is $223\ [mm]$ at rest. The hip is actuated with a brushless outrunner motor (T-motor MN-4006) with a 1:5 gearbox. The motor control board (Texas Instruments TMS320F28069M with DRV8305EVM booster packs) uses field-oriented control for direct torque control of the motor. A Xenomai real-time linux operating system handles high-level control. Electric power and computational power are both off-loaded via tether. A representative mass is directly attached on the boom just behind the leg. With the entire payload, the robot has a body weight of $600\ [g]$. For our two training wheel environments the representative mass is replaced with an intermediate mass or completely removed. This results in a body weight of $505\ [g]$ ($0.84\ \g$) and $415\ [g]$ ($0.69\ \g$) respectively.
\section{RESULTS}
We test three environments: the robot with full payload and two environments with training wheels which reduce the weight to $0.69\ \g$ and $0.84\ \g$, where $\g$ is the weight of the robot in the original environment. We will refer to these two environments with training wheels as the \emph{beginner environment} and the \emph{intermediate environment} respectively. We map out the entire reward landscape for each environment by sampling and then interpolating the parameter space (Fig. \ref{fig:3landscapes}). 
The parameter space is limited to $\theta_0 = \left[0\ 40\right]$\textdegree{} and $\theta_1 = \left[10\ 45\right]$\textdegree{}. Parameters outside this range are either unreachable due to mechanical hard-stops, or in the zero-reward region for all environments, and cropped for clarity. \\
All three landscapes have a mountain-like shape emerging out of a flat surface. While not quite convex, the landscapes each have a prominent peak, making them amenable to stochastic gradient descent. Also present in all three landscapes is a cliff: a sudden sharp drop from high to zero reward. This is found in the upper right quadrant of the parameter space and can be recognized in Fig. \ref{fig:3landscapes} by the dense contour lines. This cliff represents the border between parameters which exhibit stable high performance and unstable parameters. In practice it is both difficult as well as dangerous to learn from beyond this cliff: policies with high-amplitude tend to crash violently and damage the hardware. It is interesting to note that the orientation of the cliff does change in each environment, though its proximity to the peak does not.
\subsection{Salient Gradient Sets}
We are interested in the region that achieves non-zero reward which we will refer to as a \emph{salient gradient set} (SCS), delimited in the figures by the thicker, outermost contour. This is the set of parameters a learning agent needs to sample from in order to learn. The second criterion for choosing effective training wheels is \emph{how much the training wheels increase the size of the SCS}. Indeed the SCS of the beginner environment covers 46\% of the total parameter space, compared to 25\% of the intermediate environment and 20\% in the original environment. This increase in size is important both for gradient-based and gradient-free methods.\\
With stochastic gradient descent, for example, the gradient is estimated by local sampling. This means the agent must start inside, or at least within local sampling distance of the SCS in order to estimate a gradient. Increasing the size of the SCS directly increases the basin of attraction of the learning system. Other exploration approaches such as eps-greedy can sample from the SCS despite being initialized well away of the SCS. In this case, increasing the size of the SCS increases the probability of sampling from it, thus improving rate of convergence.

\subsection{Funneling Sets}
In Randl{\o}v's simulated work \cite{randlov2000shaping}, the training-wheels environment converges gradually to the original environment. In practice, it is often difficult to implement a gradual mechanical change: in many cases it is desirable to have training wheels that are either on or off, or at least require only a few stages. This brings up an important requirement for training wheels: \emph{successive environments must funnel into each other}. 
\begin{figure}[bthp]
    \centering
    \includegraphics[width=0.9\columnwidth]{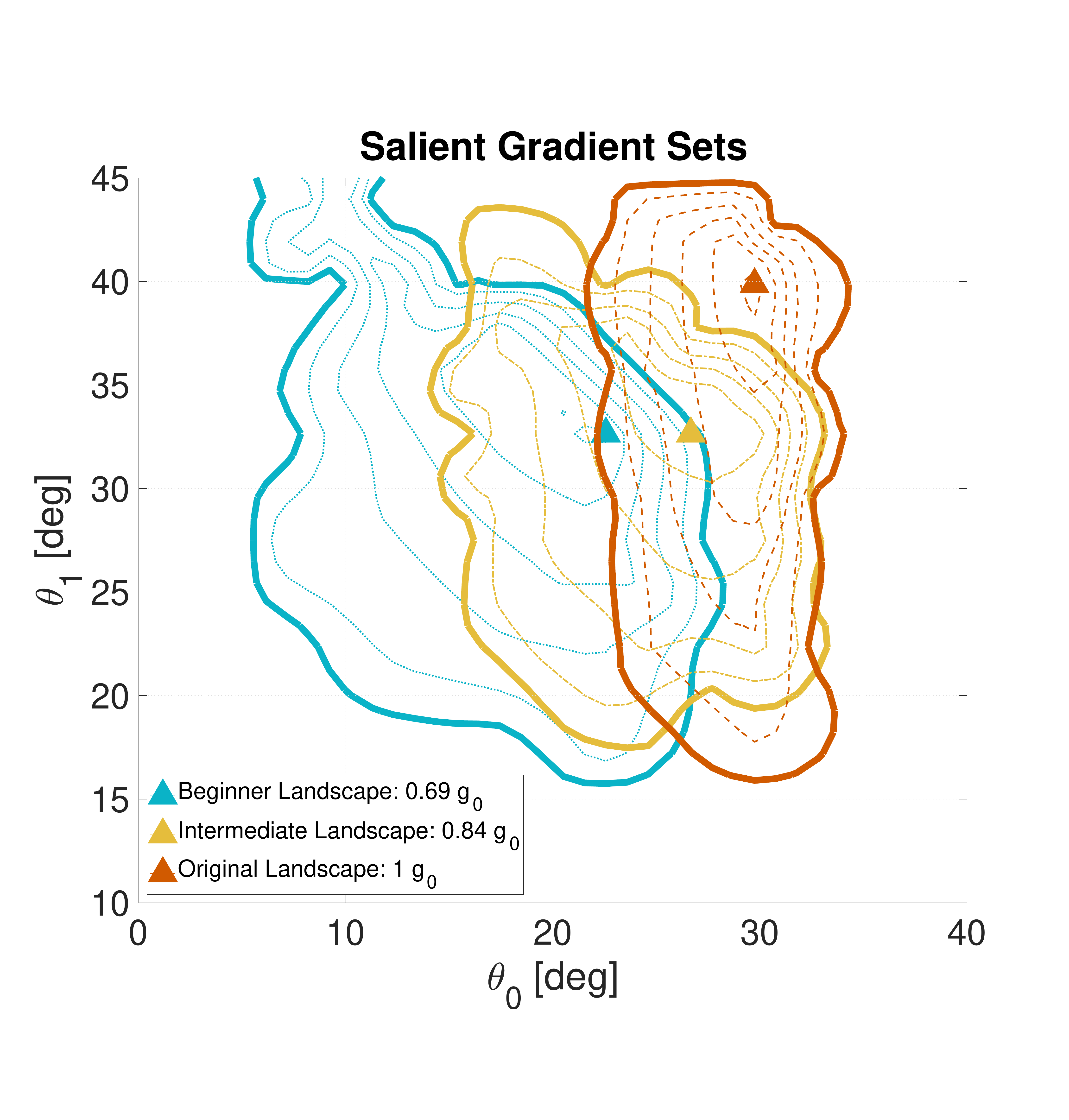}
    \caption{The salient gradient set (SCS) for each environment is mapped out with contour lines and the peak of each set marked by a triangle. The location of the peak of one training environment with respect to the SCS of the successive environment is very important. To be effective, the training wheels must guide the current policy towards parameters that will sample from the salient gradient set of the next landscape with higher probability.}
    \label{fig:salientpeaks}
\end{figure}
In Fig. \ref{fig:salientpeaks}, the peak of the beginner environment, located at $[22.5\ 32.6]$\textdegree{}, lies only barely within the SCS of the original environment. In general, there are no guarantees the peak of the training-wheels will be contained in the SCS of the original problem. If it isn't, or in our case if the policy has not fully converged, there is a good chance that when the training wheels are removed, the policy is still too far from the next SCS to effectively sample from it. This issue is solved by having an intermediate environment, whose SCS contains a large area around the peak of the beginner environment. In other words, each training-wheel environment should easily \emph{funnel into} the SCS of the next environment to be effective. This is particularly important when local exploration strategy is used, and is conceptually similar to designing controller funnels \cite{cao2015control}\cite{tedrake2010lqr}. \\
There is a second consideration that should be kept in mind: while the peak of an earlier environment \emph{must}\footnote{This condition is necessary when exploration is strictly local, and can be relaxed otherwise.} be contained in the SCS of the successive environment, the reverse is not true. This means switching \emph{back} to an earlier training environment must be done only cautiously, especially in hardware. A simple workaround is to keep a memory of previous policies, and switch back to known stable policies when switching back to previous training environments.
\subsection{LEARNING ACROSS LANDSCAPES}
\begin{figure*}[htp]
\subfloat[Typical successful learning paths]{%
  \includegraphics[clip,width=\columnwidth]{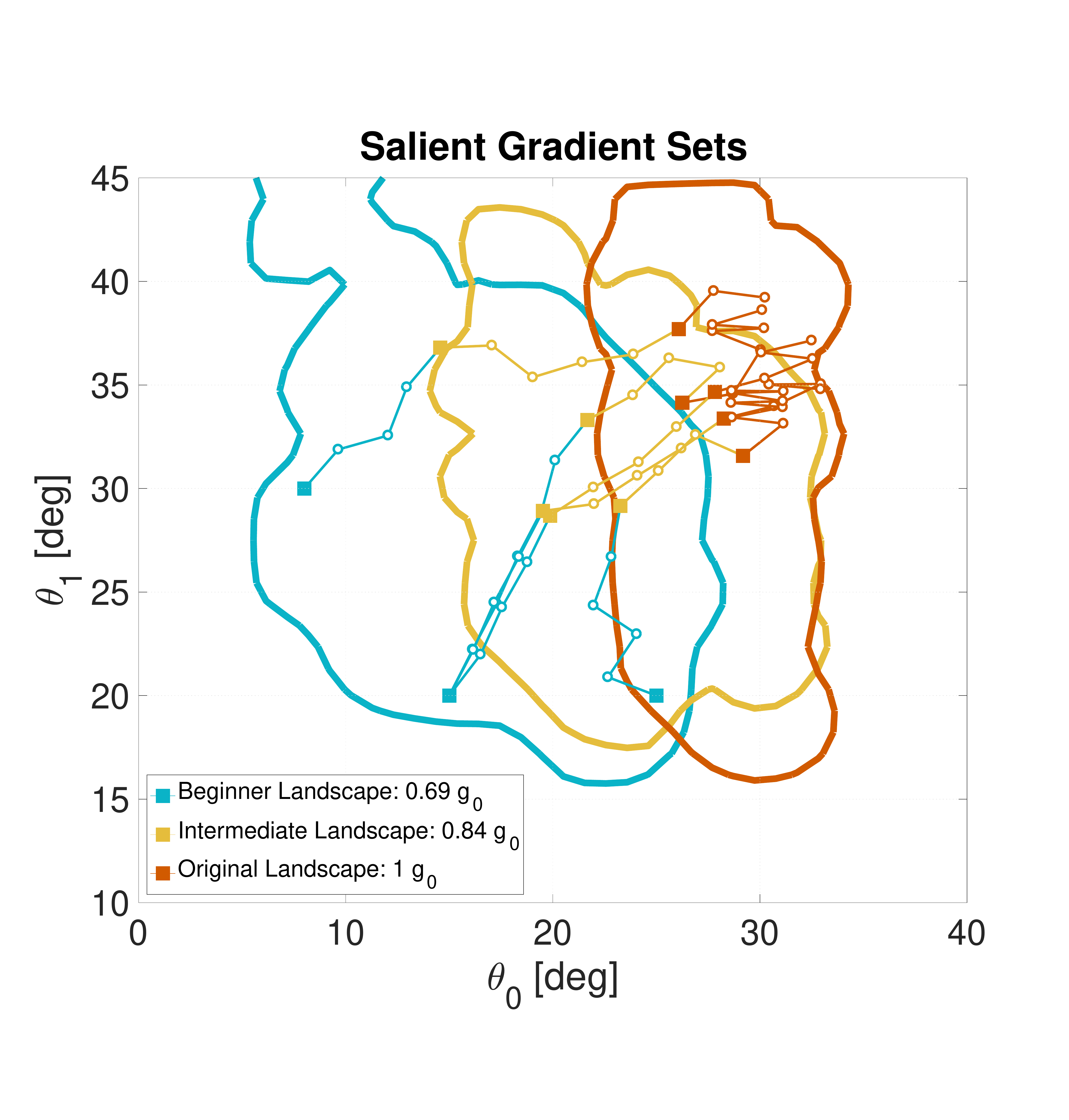}%
  \label{fig:learning}
} 
\subfloat[Unsuccessful learning attempted with the original landscape, marked in red.]{%
  \includegraphics[clip,width=\columnwidth]{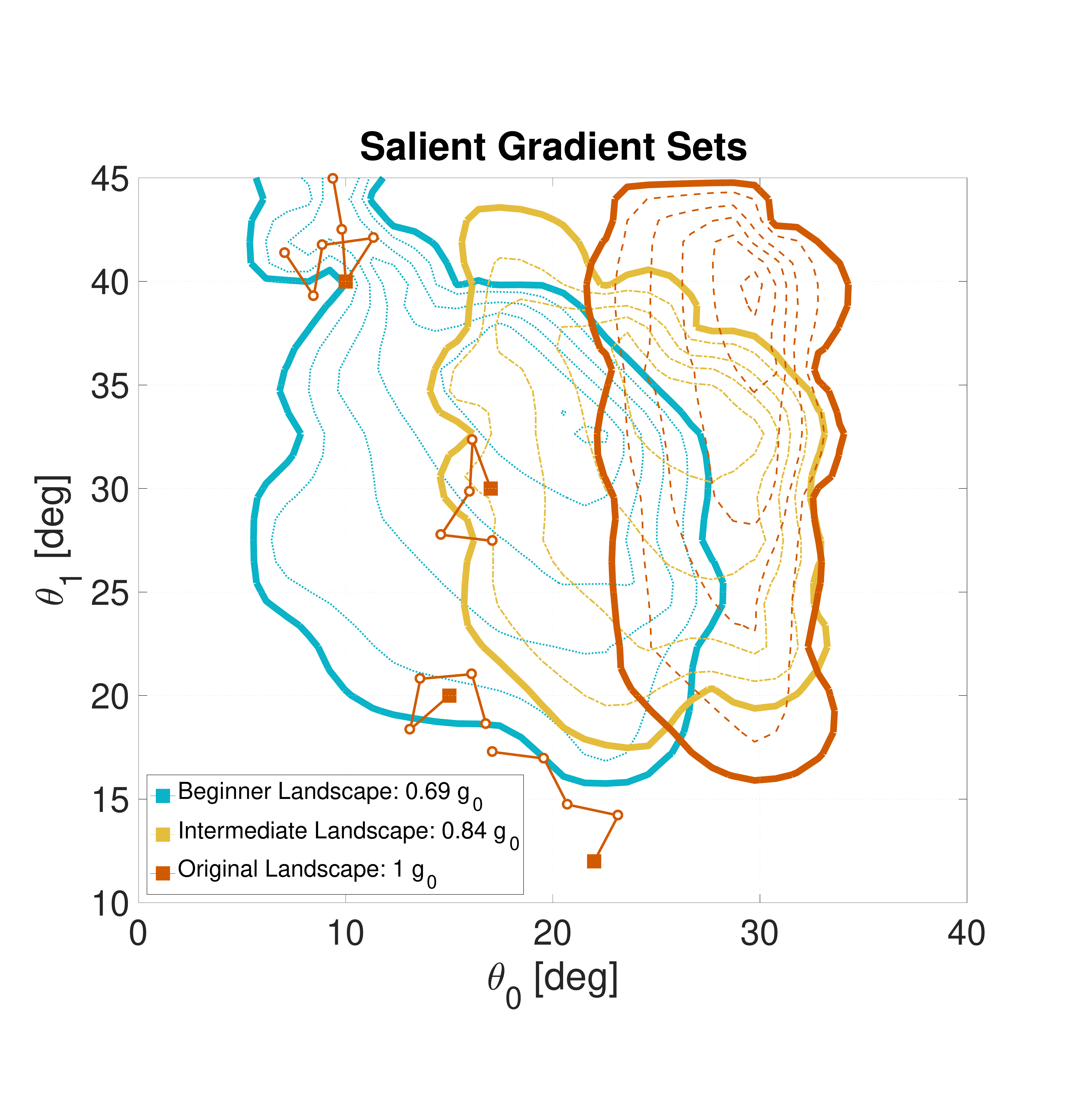}%
  \label{fig:fails}
}
\caption{Several typical learning paths are shown here: above, successful learning trials progress from the beginner to an intermediate and finally to the original environment. Contour lines are not shown for visual clarity. One of these learning trials typically took around 10 minutes. Below are trials initialized directly with the original environment but outside its salient gradient set. These take learning steps in random directions without improving.}
\end{figure*}

As a proof of concept we use offline stochastic gradient descent with finite-differences, with parameter perturbations ranging between $0.5$\textdegree{} and $2$\textdegree{}. Gradient estimates tend to be more robust to noise with larger perturbations, especially where the gradient is very shallow such as at the edges of the SCS. On the other hand, they become unstable closer to the peak and especially when close to the cliff. We also choose a constant, relatively large learning rate of $2.5$. Again, larger steps have the risk of overshooting and stepping over the cliff, but otherwise perform well. In both cases a cleverer, variable choice of these parameters would help the learning process, but is not relevant to the training-wheels concept and is kept constant. \\
Several typical learning sessions are shown in Fig. \ref{fig:learning}, with the most successful reaching a speed of $0.35\ [\frac{m}{s}]$. We also purposefully initialize several trials outside the SCS of the original environment, and as expected we observe meandering paths.
Examples of agents initialized with parameters outside their respective SCS are shown in Fig. \ref{fig:fails}. As expected, without sufficient gradient information the agents will simply take steps in random direction. While this random exploration has a non zero probability of entering the salient gradient set and therefore converging, it can take a large amount of iterations, especially when starting at some distance from the set and with smaller learning step sizes. Especially when learning directly in robot hardware, it is critical to reduce the number of trials necessary.


\section{CONCLUSIONS AND OUTLOOK}

We build on the concept of training wheels, temporary mechanical modifications of the system, to shape the learning landscape \cite{randlov2000shaping}. We apply it to learning open-loop legged locomotion in a constrained test stand, as a simple, low-dimensional problem that is unstable, underactuated and features impacts. We propose three criteria to designing effective training wheels in practice.
\begin{enumerate}
    \item Ease of application to a generic set of robots
    \item Increase in probability of sampling from the salient gradient set
    \item Ease of funneling from a training environment to the successive environment
\end{enumerate}
Since reducing the engineering effort is a main attraction for applying learning to robotics, it is important that training wheels are easy to implement and apply. As an example, we surmise that adding damping to the joints, or to the floating base, of a robot would help stabilize the system and greatly help learning by improving stability \cite{colgate1994factors}\cite{calisti2015dynamics}, at a moderate cost to performance and efficiency. Implementing mechanical damping on small joints is however much more difficult than simply temporarily offloading some of the payload, and would require a custom design for each new robot. This partially defeats the purpose of reducing the engineering effort. While we plan to explore solutions to this in future work, there is a lot of merit in solutions as effective yet simple as reducing the payload. \\
The second criterion is the main qualifier for the effectiveness of the training wheels in shaping the learning landscape. To be more precise, increasing the probability of sampling from the salient gradient set is what makes a training wheel environment easier to learn in. The actual size of the set in relation to the sample space is a good proxy; it is more generalizable to arbitrary exploration strategies, and makes it more intuitive to predict when designing training wheels. In most cases it will be impossible to map out the landscape by brute-force as we have done here. Good methods to approximate the SGS size, or directly estimate the sampling probability, need to be developed to more systematically evaluate potential training wheels. \\
The final criterion is particularly relevant when local exploration strategies are used. As this strategy is common in robotics, we feel it is an important criterion to include. Training wheels that can be continuously tuned out, until the dynamics converge back to the original environment, would be guaranteed to satisfy this criterion \cite{randlov2000shaping}. However the implementation of such training wheels generally goes against the first criteria, and a trade-off will have to be made. To be effective, training wheels will require a strategy to transition from one landscape to the next. In practice, it is helpful to regularly estimate the local gradient of the next landscape before each transition. Ideally, we would want to design a sequence of training wheels that funnel into each other, similar in concept to \cite{cao2015control}\cite{karpathy2012curriculum}. \\
In this work, when to switch between environments was chosen heuristically. With the actual landscape maps available for reference, we can be very confident that the funnel overlap between environments is large and we do not need to completely converge on one environment before switching to the next. For future work, it will be interesting to find a more general rule for switching environments. Since the number of trials needed to converge is particularly important when learning in hardware, an optimal switching policy to learn with the fewest iterations would be particularly useful. \\
Although we have presented these landscape shaping results in the context of reinforcement learning, the challenge of traversing a landscape in parameter-space is inherent to optimization problems as a whole. In particular, the concepts we develop should be useful for applying derivative-free optimization in hardware \cite{sproewitz2008learning} as well.





\section*{ACKNOWLEDGMENTS}
We would like to thank {\"O}zge Drama, Felix Grimminger and Julian Viereck for fruitful discussions, advice and help along the way. We also thank Felix Widmaier who wrote the code to run and interface with the motor control boards. We thank the International Max Planck Research School for Intelligent Systems (IMPRS-IS) for supporting the academic development of Felix Ruppert. This work was partially supported by an MPI Grassroots grant provided to Ludovic Righetti, Felix Grimminger and Alexander Spr{\"o}witz in 2017.\\


\printbibliography

\end{document}